\documentclass[onecolumn]{fairmeta}

\usepackage{graphicx}
\usepackage{booktabs}
\usepackage{multirow}
\usepackage{wrapfig}
\usepackage{enumitem}
\usepackage{amsfonts} 
\usepackage{amsmath}
\usepackage{subcaption}
\usepackage{graphicx}
\usepackage{wrapfig}
\usepackage{multicol}
\usepackage{colortbl}
\usepackage{tabularx}
\usepackage{fontawesome5}

\newcommand{\fref}[1]{Fig.~\ref{#1}}
\newcommand{\sref}[1]{Sec.~\ref{#1}}
\newcommand{\tref}[1]{Table~\ref{#1}}
\newcommand{\aref}[1]{Appendix~\ref{#1}}
\newcommand{\methodname}{\textsc{PRISM}}
\newcommand{\RQone}{What evidence supports implicit spatial representations for robotic manipulation?}
\newcommand{\RQtwo}{Does spatial compression limit geometry-sensitive manipulation?}
\newcommand{\RQthree}{How does {\methodname} improve low-resolution manipulation performance?}
\newcommand{\RQfour}{How does each component of {\methodname} contribute to performance?}
\newcommand{\RQfive}{Can {\methodname} be deployed in multi-task and real-world scenarios?}

\newcommand{\sspool}{SSPool}
\newcommand{\avgpool}{AvgPool}
\newcommand{\maxpool}{MaxPool}
\newcommand{\nopool}{NoPool}

\definecolor{rowgray}{gray}{0.92}
\definecolor{cellgray}{gray}{0.94}
\newcolumntype{Y}{>{\centering\arraybackslash}X}

\title{Rethinking Implicit Spatial Representation in Visuomotor Policy Learning}

\author[1]{\href{https://xyc0212.github.io/}{\textcolor{black}{Xiangyu Chen}}}
\author[1]{Yuxuan Hu}
\author[1]{Chuhao Zhou}
\author[1,\dagger]{\href{https://marsyang.site/}{\textcolor{black}{Jianfei Yang}}}
\affiliation[1]{MARS Lab, Nanyang Technological University}
\affiliation[\dagger]{Corresponding Author}
\correspondence{\email{jianfei.yang@ntu.edu.sg}.}

\abstract{
Generative model-based imitation learning has become a widely adopted paradigm for robotic manipulation, where policy performance depends critically on the conditioned visual representations.
Although spatial softmax-based representations have been adopted in prior visuomotor policies, their effectiveness and underlying mechanisms remain insufficiently understood.
This work rethinks the use of spatial softmax pooling: do such implicit spatial representations provide effective and stable visual features for robotic manipulation?
Through systematic studies of different pooling methods in visual encoders, we find that this pooling operation produces compact and stable spatial representations, which outperform feature-value representations, despite using substantially fewer dimensions.
Complementary saliency analysis further suggests that these spatial representations guide the encoder to focus more consistently on task-relevant regions.
However, this advantage is limited by a representation bottleneck in current visual encoders: repeated downsampling operations weaken fine-grained spatial information before the action-generation module can use it, especially under low-resolution observations.
Motivated by these findings, we propose {\methodname}, a visual encoder that preserves multiscale implicit spatial information through top-down cross-attention fusion.
Experiments across multiple tasks and policy backbones show consistent improvements. 
In particular, on the low-resolution, high-precision \textit{ToolHang} task, {\methodname} shows clear gains, improving the average success rate from $5.0\%$ to $13.4\%$ while increasing parameters by only $15.4\%$. 
These results support the use of multiscale implicit spatial representations as an effective and efficient design principle for robotic manipulation.
}

\begin{document}

\maketitle


\section{Introduction}

Imitation learning~\citep{osa2018algorithmic} has become a widely adopted paradigm for robotic manipulation, enabling robots to acquire complex manipulation skills directly from human demonstrations.
Recent generative model-based policies~\citep{chi2025diffusion, zhan2026mean} have demonstrated strong generalization capability by denoising Gaussian noise into action distributions conditioned on visual features.
These approaches capture multimodal behaviors and show stronger robustness, substantially advancing robotic manipulation.

Visual representation is therefore central to the generative model-based policies.
Most existing policies adopt standard visual encoders, such as ResNet backbones~\citep{he2016deep}, to extract compact representations from images.
These encoders largely inherit design choices from semantic recognition, where repeated spatial downsampling and global aggregation are effective for capturing object categories and appearance cues.
As a result, the learned representation tends to emphasize what visual patterns are present, rather than preserving where they occur and how they are spatially arranged~\citep{tomasini2023deep}.

Unlike semantic vision tasks, manipulation requires precise spatial cues for action prediction, 
including object pose, contact structure, and robot-object alignment.
Spatial representations therefore provide a natural alternative to globally aggregated semantic feature representations, 
as they expose relative position information to visuomotor policies.
Although spatial softmax pooling ({\sspool})~\citep{finn2016deep} has been adopted in prior visuomotor policies~\citep{chi2025diffusion, mandlekar2022matters}, 
it is often treated as an architectural component rather than as a representation mechanism.
Its role and effectiveness remain underexplored.
Thus, our work rethinks the use of {\sspool} and begins with the question:
\textbf{Do implicit spatial representations provide effective and stable visual features for robotic manipulation?}

We answer this question through controlled pooling ablations across three generative policy backbones, varying only what representations are extracted from the final feature maps.
The results show that implicit spatial representations from {\sspool} consistently improve policy performance with substantially fewer feature dimensions.
Saliency analysis further suggests that these spatial representations provide effective and stable visual conditioning for robotic manipulation.

Beyond the choice of representation, we further identify \emph{spatial information collapse} as an overlooked encoder-level bottleneck in visuomotor policy design.
Although standard visual encoders, such as ResNet-style backbones, are commonly adopted as default feature extractors, repeated downsampling and global aggregation can produce overly coarse final feature maps, weakening the fine-grained spatial structure needed for precise pose estimation and tool-object alignment.
While high-resolution images or additional camera views can alleviate this issue, they increase system complexity without directly addressing the bottleneck inside the encoder.

Motivated by these findings, we propose \emph{{\methodname}}, a visual encoder that preserves multiscale implicit spatial representations for robotic manipulation.
Rather than extracting implicit spatial representations only from the final feature map, {\methodname} extracts coordinate-aware features from multiple residual stages and fuses them through a top-down cross-attention mechanism into a compact implicit spatial representation.
This design preserves spatial cues across multiple feature resolutions, strengthens the spatial information available to the policy, and mitigates the spatial information collapse caused by aggressive encoder compression.
In summary, our contributions are as follows:
\begin{itemize}[leftmargin=*, itemsep=1pt, topsep=2pt, parsep=0pt]
    \item We provide empirical evidence and saliency analysis, demonstrating that implicit spatial representations are more effective and stable for robotic manipulation than globally aggregated feature-value representations, despite using substantially fewer feature dimensions.
    \item We identify spatial information collapse as an encoder-level bottleneck and empirically show that aggressive compression in conventional visual encoders weakens the fine-grained spatial structure required for low-resolution, geometry-sensitive manipulation.
    \item Motivated by this diagnosis, we propose \emph{{\methodname}}, a visual encoder that extracts and fuses multiscale implicit spatial representations through multiscale spatial softmax extraction and top-down cross-attention fusion, achieving consistent improvements across manipulation tasks.
\end{itemize}

\section{Related Work}
\label{sec:related}

\paragraph{Generative-Model-Based Policies}
Generative-model-based policies~\citep{chi2025diffusion, geng2026mean, black2024pi_0} have become a strong paradigm for robotic manipulation.
These methods denoise actions from pure noise, conditioned on visual observations and robot proprioception, making them suitable for modeling multimodal demonstrations.
Recent policies predict action chunks from image observations for closed-loop manipulation~\citep{zhao2023learning}.
While these works~\citep{a2a2026, an2026feedback, bai2026flash} improve action generation, the visual representation used to condition the policy remains critical.
Most existing policies~\citep{li2025mathbf} adopt ResNet backbones~\citep{he2016deep} to extract image features, although these encoders are originally developed for semantic recognition tasks such as image classification~\citep{deng2009imagenet}.
This motivates our study of which visual representations provide effective and stable conditioning signals for generative visuomotor policies.

\paragraph{Spatial-Information-Augmented Policies}
Spatial information is essential for robotic manipulation, where policies infer object pose, end-effector pose, and robot-object alignment.
Prior works incorporate spatial cues through coordinate-aware features~\citep{hu2026trace, ye2026spatially}, keypoints~\citep{fang2025kalm, zheng2025tracevla}, and object poses~\citep{an2024rgbmanip, deng2025graspvla}.
Spatial softmax pooling~\citep{chi2025diffusion, finn2016deep, mandlekar2022matters} converts convolutional feature activations into compact coordinate-like representations preserving the spatial relationship in deep features.
However, its role as an implicit spatial representation mechanism remains insufficiently studied.
Moreover, aggressive striding and pooling in image encoders collapse spatial information and produce coarse feature maps~\citep{fu2024featup}.
Motivated by this gap, we systematically study implicit spatial representations for robotic manipulation and propose {\methodname} to preserve them across multiple feature scales.

\section{Implicit Semantic vs. Implicit Spatial Representations}
\label{sec:semantic_vs_spatial_representation}

A visual encoder maps an input image to deep feature maps, while the representation head determines what information from these maps is exposed to the policy.
Conventional heads, such as average pooling ({\avgpool}) and max pooling ({\maxpool}), aggregate feature activations into value-based descriptors that capture which visual patterns are present.
In contrast, {\sspool} couples activations with a coordinate grid, producing coordinate-aware descriptors that capture where these patterns occur.
The key distinction is therefore whether the representation only summarizes activation values or also preserves the spatial locations of feature responses.
Based on this distinction, we categorize visual representations into \emph{implicit semantic representations} and \emph{implicit spatial representations}.

\paragraph{Implicit Semantic Representation.}
Conventional visual encoders often use global aggregation, such as {\avgpool} and {\maxpool}, to obtain compact feature representations.
Given a feature map $\mathbf{F}\in\mathbb{R}^{C\times H\times W}$, these operations extract a scalar value from each channel.
The resulting vector mainly reflects the presence or strength of visual features, without explicitly preserving their relative spatial locations.
We therefore refer to these representations as \emph{implicit semantic representations}.

\paragraph{Implicit Spatial Representation.}
In contrast, we refer to the output of {\sspool}~\citep{finn2016deep, levine2016end} as \emph{implicit spatial representations} since it combines feature activations with coordinate grids.
For each channel, it normalizes activations over spatial locations and computes the expected coordinate:
\begin{equation}
    p_{cij} =
    \frac{\exp(F_{cij}/\tau)}
    {\sum_{m=1}^{H}\sum_{n=1}^{W}\exp(F_{cmn}/\tau)},
    \qquad
    (x_c,y_c)=\sum_{i=1}^{H}\sum_{j=1}^{W}p_{cij}(x_{ij},y_{ij}),
\end{equation}
where $\tau$ is a temperature parameter and $(x_{ij},y_{ij}) \in [-1,1]^2$ denotes the normalized coordinate of location $(i,j)$.
Unlike {\avgpool} and {\maxpool}, which summarize activation values, {\sspool} converts each channel into a spatial distribution and outputs the expected coordinate of the corresponding feature response.
This provides coordinate-aware visual cues that are directly relevant to pose, alignment, and contact reasoning in robotic manipulation.
This distinction provides the basis for our experiments, where we compare implicit semantic and implicit spatial representations.

\section{Methodology}
\label{sec:method}
\begin{wrapfigure}[14]{r}{0.44\linewidth}
    \centering
    \vspace{-14mm}
    \includegraphics[width=0.96\linewidth]{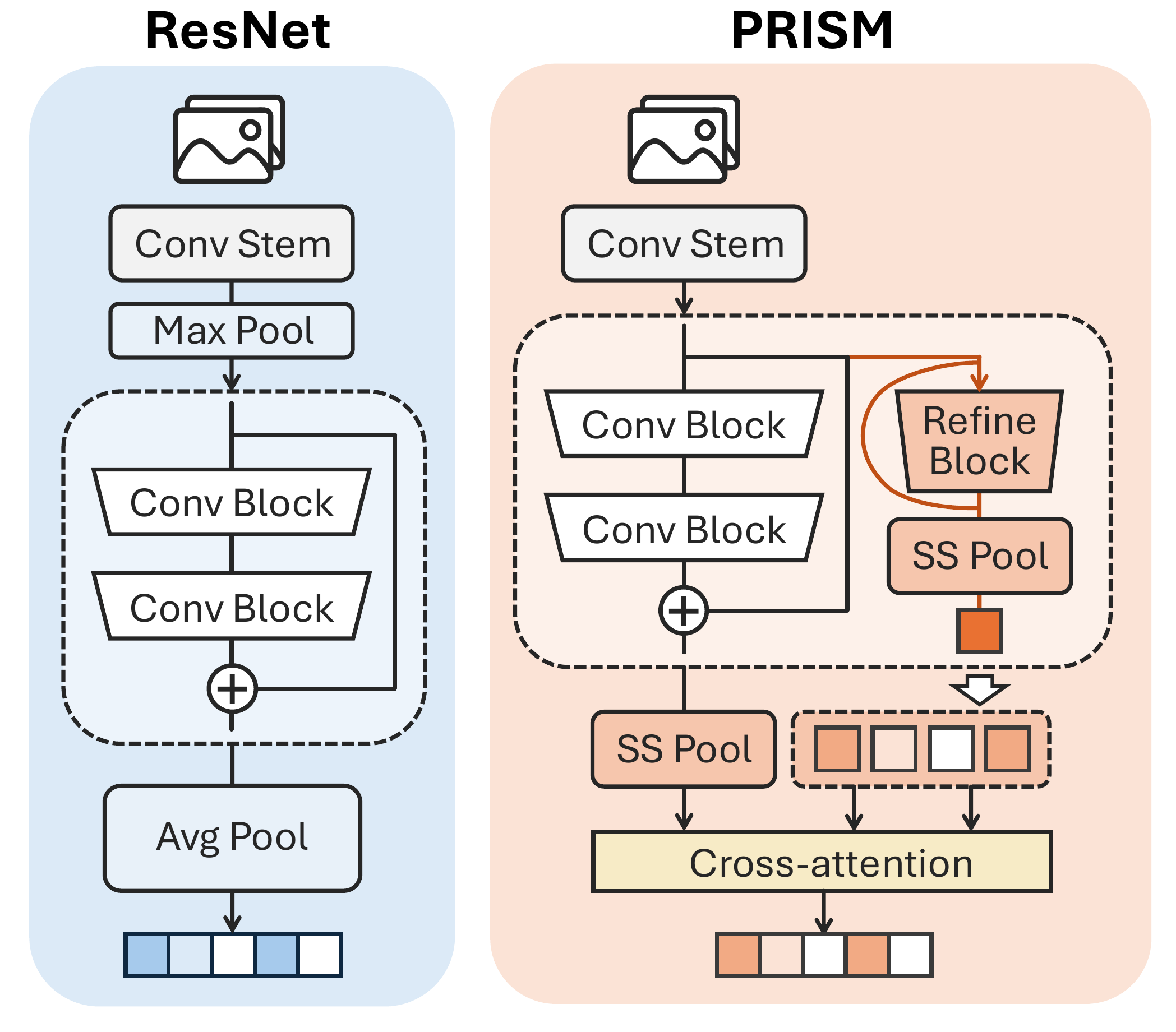}
    \vspace{-0mm}
    \caption{\textbf{Overview of {\methodname}.}}
    \label{fig:methodology}
\end{wrapfigure}
This section introduces {\methodname}, a visual encoding framework for preserving multiscale implicit spatial representations in robotic manipulation.
While {\sspool} provides an effective coordinate-aware representation, applying it only to the final feature map can be limiting because current encoders repeatedly downsample visual features before pooling.
This compression may weaken fine-grained spatial cues before they are exposed to the policy.
To address this, {\methodname} extracts implicit spatial representations from different residual stages using multiscale spatial softmax (MS-SS) and aggregates these multiscale cues into a compact policy-conditioning vector through top-down cross-attention~(TDCA), as shown in~\fref{fig:methodology}.

\subsection{Multiscale Implicit Spatial Softmax}
In standard ResNet-based encoders, {\sspool} is commonly applied only after multiple downsampling stages as the replacement of average pooling.
While effective for recognition-oriented perception, this final-stage pooling can be too compressed for manipulation tasks requiring precise pose estimation, alignment, and contact reasoning.
Therefore, {\methodname} extracts implicit spatial representations from multiple residual stages, as illustrated by the right part in~\fref{fig:methodology}.
Specifically, given the feature map $\mathbf{x}^{l}$ from the $l$-th stage, we apply a residual refinement block before {\sspool}:
\begin{equation}
    \hat{\mathbf{x}}^{l}
    =
    \mathbf{x}^{l} + \mathcal{F}_{\mathrm{ref}}^{l}(\mathbf{x}^{l}),
    \qquad
    \mathbf{s}^{l} = \mathrm{SSPool}^{l}(\hat{\mathbf{x}}^{l}),
\end{equation}
where $\mathcal{F}_{\mathrm{ref}}^{l}(\cdot)$ is the refine block, $\hat{\mathbf{x}}^{l}$ is the refined feature map, and $\mathbf{s}^{l}$ is the extracted implicit spatial representation.
The multiscale spatial representations $\{\mathbf{s}^{l}\}_{l=1}^{L}$ are passed to the fusion module.

\subsection{Top-Down Cross-Attention Fusion}
MS-SS provides complementary multiscale implicit spatial representations.
The deepest representation contains stronger task-level context due to its larger receptive field, while earlier representations retain finer spatial details with less compression.
Direct concatenation combines these representations but treats all scales equally.
To enable high-level context to select relevant fine-scale cues, {\methodname} uses a lightweight top-down cross-attention (TDCA) fusion module.

Given the multiscale spatial representations $\{\mathbf{s}^{l}\}_{l=1}^{L}$ from MS-SS module, we use the deepest spatial representation $\mathbf{s}^{L}$ as the query and the earlier-stage representations $\{\mathbf{s}^{l}\}_{l=1}^{L-1}$ as the keys and values.
For compact notation, we denote the concatenation of earlier-stage representations as
$\mathbf{s}^{1:L-1} = [\mathbf{s}^{1}; \mathbf{s}^{2};\ldots; \mathbf{s}^{L-1}]$.
Therefore, the TDCA is formulated as
\begin{equation}
    \mathbf{z}
    =
    \mathrm{softmax}
    \left(
    \frac{Q(\mathbf{s}^{L})K(\mathbf{s}^{1:L-1})^\top}{\sqrt{d}}
    \right)
    V(\mathbf{s}^{1:L-1}),
\end{equation}
where $Q(\cdot)$, $K(\cdot)$, and $V(\cdot)$ are learnable projections, and $d$ is the attention dimension.
The attended feature $\mathbf{z}$ is then projected to a fixed-dimensional vector as the visual condition for the downstream policy.
By using the deepest spatial representation as high-level context, TDCA retrieves relevant finer-scale spatial cues from earlier stages.
This allows {\methodname} to efficiently aggregate multiscale implicit spatial representations while keeping the policy-conditioning vector compact.

\section{Experimental Results}
\label{sec:experiments}

The central contribution of this work is empirical: we study whether implicit spatial representations provide effective visual features for robotic manipulation and whether {\methodname} can strengthen them. 
Our experiments are organized around five research questions:
\begin{description}[
    leftmargin=1.2cm,
    labelwidth=0.9cm,
    labelsep=0.3cm,
    itemsep=1pt,
    topsep=2pt,
    font=\bfseries
]
    \item[RQ1.] \textbf{\RQone}
    \item[RQ2.] \textbf{\RQtwo}
    \item[RQ3.] \textbf{\RQthree}
    \item[RQ4.] \textbf{\RQfour}
    \item[RQ5.] \textbf{\RQfive}
\end{description}

We answer these questions through pooling ablations, bottleneck analysis on high-precision tasks, comparisons with ResNet-based baselines, and component ablations. 
We further evaluate {\methodname} on LIBERO~\citep{liu2023libero} multi-task benchmarks to examine generalization beyond single-task settings, and include real-world experiments as physical deployment validation. All encoder comparisons use the same policy architectures, training schedules, and evaluation protocols to isolate the effect of visual representation. Detailed experimental settings are provided in~\aref{app:training_details}.

\subsection{\RQone}
\label{sec:rq1}

\begin{table}[t]
\centering
\caption{
\textbf{Pooling ablation under standard and truncated ResNet18 backbones.}
Results are reported as mean success rate (\%) across the three short-horizon robomimic tasks with three seeds.
}
\label{tab:pooling_ablation_parallel_delta}
\small
\setlength{\tabcolsep}{3.0pt}
\renewcommand{\arraystretch}{1.08}

\begin{tabular}{lccccc@{\hspace{6pt}}cccc}
\toprule
& \multicolumn{5}{c}{\textbf{Standard ResNet18}}
& \multicolumn{4}{c}{\textbf{Truncated ResNet18}} \\
\cmidrule(lr){2-6}
\cmidrule(lr){7-10}
\textbf{Pooling}
& \textbf{Dim.}
& \textbf{DP}
& \textbf{FM}
& \textbf{MF}
& \textbf{Avg.}
& \textbf{DP}
& \textbf{FM}
& \textbf{MF}
& \textbf{Avg.} \\
\midrule

\rowcolor{cellgray}
\textbf{\sspool}
& 64
& \textbf{89.7}
& \textbf{88.7}
& \textbf{91.6}
& \textbf{90.0}
& \textbf{87.3} {\scriptsize (-2.4)}
& \textbf{91.3} {\scriptsize (+2.6)}
& \textbf{90.7} {\scriptsize (-0.9)}
& \textbf{89.8} {\scriptsize (-0.2)} \\

\avgpool
& 512
& 70.9
& 86.0
& 86.4
& 81.1
& 61.6 {\scriptsize (-9.3)}
& 87.1 {\scriptsize (+1.1)}
& 84.2 {\scriptsize (-2.2)}
& 77.6 {\scriptsize (-3.5)} \\

\maxpool
& 512
& 68.4
& 75.1
& 72.2
& 71.9
& 67.1 {\scriptsize (-1.3)}
& 72.4 {\scriptsize (-2.7)}
& 66.0 {\scriptsize (-6.2)}
& 68.5 {\scriptsize (-3.4)} \\

\nopool
& 4608
& 68.2
& 78.2
& 65.8
& 70.7
& 67.1 {\scriptsize (-1.1)}
& 78.4 {\scriptsize (+0.2)}
& 73.1 {\scriptsize (+7.3)}
& 72.9 {\scriptsize (+2.2)} \\
\bottomrule
\end{tabular}
\vspace{-7mm}
\end{table}

To answer this question, we isolate the effect of representation type by comparing policies with the same ResNet~\citep{he2016deep} encoder but different pooling strategies: average pooling (\avgpool), max pooling (\maxpool), no pooling (\nopool), and spatial softmax pooling (\sspool).
While {\avgpool}, {\maxpool}, and {\nopool} aggregate feature activations as semantic feature-value representations, {\sspool} transforms feature activations into coordinate-like representations by computing channel-wise implicit spatial locations.
This comparison allows us to directly examine which implicit representations provide a more effective visual conditioning signal for robotic manipulation.

\begin{wrapfigure}[12]{r}{0.44\linewidth}
    \centering
    \vspace{-6mm}
    \includegraphics[width=\linewidth]{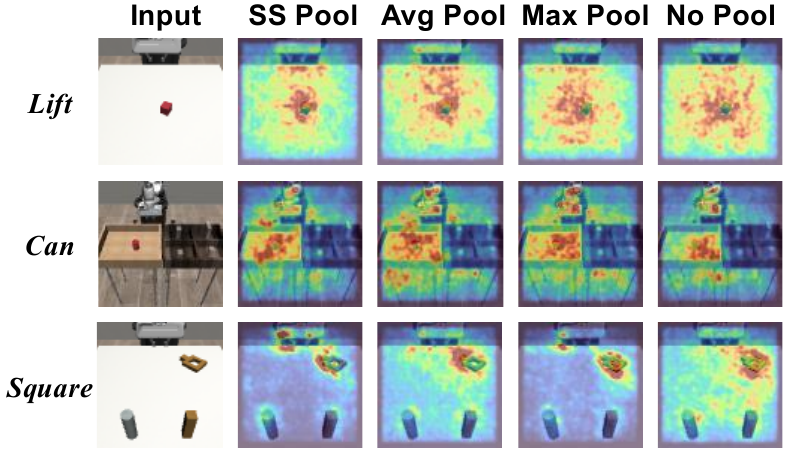}
    \caption{
    \textbf{Saliency Maps of different pooling strategies for robomimic tasks.}
    }
    \label{fig:saliency_maps_pool_3_tasks}
\end{wrapfigure}

We evaluate the four pooling strategies with Diffusion Policy (DP)~\citep{chi2025diffusion, ho2020denoising}, Flow Matching (FM)~\citep{lipman2023flow}, and MeanFlow (MF)~\citep{zhan2026mean,geng2026mean} on \textit{Lift}, \textit{Can}, and \textit{Square}, as shown in the first column of~\fref{fig:saliency_maps_pool_3_tasks}.
Given $84\times84$ observations, the standard ResNet18 encoder outputs a $512\times3\times3$ feature map.
{\sspool} maps it to 64 coordinate features, whereas average and max pooling produce 512-dimensional descriptors and no pooling yields a 4608-dimensional flattened feature.

First, we visualize saliency maps for the three short-horizon tasks in~\fref{fig:saliency_maps_pool_3_tasks} and~\aref{app:saliency_map}. 
Although all methods use the same ResNet as encoder, different pooling methods lead to clearly different attention patterns. 
{\sspool} produces the cleanest and most localized responses, concentrating on task-relevant regions such as the objects. 
In contrast, {\avgpool}, {\maxpool}, and {\nopool} produce more diffuse or less consistent responses. 
This qualitative evidence suggests that coordinate-aware {\sspool} helps the encoder preserve spatial cues that are useful for visuomotor control.

Second, we report the success rate under the same training setting with three seeds in~\tref{tab:pooling_ablation_parallel_delta}. As shown in~\tref{tab:pooling_ablation_parallel_delta}, {\sspool} achieves the best average success rate under the standard ResNet18 backbone, outperforming {\avgpool}, {\maxpool}, and {\nopool} by $8.9$, $18.1$, and $19.3$ points, respectively, with an $8\times$ to $72\times$ smaller representation.
The poor performance of {\nopool}, despite retaining the most features, shows that representation structure matters more than feature dimensionality.
Together with the saliency maps, these results suggest that clean and localized spatial responses are associated with stronger robotic manipulation performance.

\begin{wraptable}{r}{0.44\linewidth}
\centering
\vspace{-5mm}
\caption{
\textbf{Success rates (\%) of channel-shuffle ablation on \sspool.}
}
\vspace{-2mm}
\label{tab:ss_shuffle_results}
\small
\setlength{\tabcolsep}{4.2pt}
\renewcommand{\arraystretch}{0.9}
\begin{tabular}{lcccc}
\toprule
\textbf{Model} & \textbf{\textit{Lift}} & \textbf{\textit{Can}} & \textbf{\textit{Square}} & \textbf{Avg.} \\
\midrule
DP 
& 77{\textpm}6  
& 2{\textpm}2 
& 23{\textpm}6 
& 34.0{\scriptsize\,(-55.7)} \\
FM    
& 65{\textpm}29 
& 3{\textpm}2 
& 9{\textpm}7  
& 25.8{\scriptsize\,(-62.9)} \\
MF  
& 65{\textpm}8  
& 2{\textpm}0 
& 51{\textpm}8 
& 39.3{\scriptsize\,(-52.3)} \\
\midrule
\textbf{Avg.} 
& 69.0 
& 2.3 
& 27.7 
& 33.0{\scriptsize\,(-56.3)} \\
\bottomrule
\end{tabular}
\vspace{-4.5mm}
\end{wraptable}
We further repeat the comparison with a truncated ResNet18, where the final convolutional stage is removed to reduce spatial compression.
{\sspool} remains the best-performing strategy and is nearly unchanged on average ($90.0\!\rightarrow\!89.8$), whereas {\avgpool} and {\maxpool} drop by $-3.5$ and $-3.4$ points.
These results suggest that {\sspool} provides a stable spatial representation, and that the less-compressed feature maps still contain rich implicit spatial representation.
This motivates {\methodname} to preserve spatial representations across multiple feature scales.

To further examine whether {\sspool} provides structured spatial information, 
we conduct a channel-shuffle ablation.
We randomly shuffle the $K$ coordinate channels while preserving each $(x,y)$ pair during training and inference.
As shown in~\tref{tab:ss_shuffle_results}, the shuffled representation still achieves non-zero success rates, 
suggesting that the retained coordinates preserve useful spatial information.
However, the substantial performance drop shows that the original channel organization is also important.
Together, these results suggest that {\sspool} provides not only coordinate-level spatial cues, 
but also a channel-structured implicit spatial representation for policy learning.

Overall, these results show that the advantage of {\sspool} comes from compact, channel-structured spatial information, rather than from higher feature dimensionality or a specific policy formulation. 
This motivates {\methodname} to preserve such spatial structure across multiple feature scales. Thus, we use the ResNet+{\sspool} as the baseline~(Base.) in the following experiments.

\subsection{\RQtwo}
\label{sec:rq2}
\begin{wraptable}[10]{r}{0.28\linewidth}
\vspace{-5mm}
\centering
\caption{\textbf{Pooling ablation on low-res. \textit{ToolHang}.}}
\label{tab:rq2_pool_ablation}
\vspace{-2mm}
\small
\setlength{\tabcolsep}{4.2pt}
\renewcommand{\arraystretch}{0.95}
\begin{tabular}{@{}lcc@{}}
\toprule
\textbf{Pooling} & \textbf{Dim.} & \textbf{SR} \\
\midrule
\avgpool          & 512  & 7.3 \\
\maxpool          & 512  & 8.0 \\
\nopool           & 4608 & 8.7 \\
\midrule
\sspool           & 64   & 4.7 \\
\sspool~(Trunc.)  & 64   & 7.3 \\
\textbf{MS-\sspool} & \textbf{384} & \textbf{11.3} \\
\bottomrule
\end{tabular}
\end{wraptable}
To answer this question, we use the \textit{ToolHang} task with low-resolution $84{\times}84$ observations as a stress test.
Unlike the short-horizon tasks in Sec.~5.1, \textit{ToolHang} is a high-precision manipulation task that requires accurate pose estimation, tool-object alignment, and contact reasoning.
As shown in~\tref{tab:rq2_pool_ablation}, applying {\sspool} only to the final ResNet feature map achieves lower performance than the other pooling choices, while {\nopool} performs best among the baseline representations.
This discrepancy with the results in~\sref{sec:rq1} suggests a possible bottleneck: when low-resolution observations are used for geometry-sensitive manipulation, the aggressively compressed final feature map may be too coarse for {\sspool} to extract sufficiently fine-grained implicit spatial information.

To test this hypothesis, we provide {\sspool} with less-compressed and multiscale spatial features.
Using TruncResNet, which removes the final convolutional stage and preserves a less-compressed feature map, improves the success rate by $55\%$.
Moreover, multiscale {\sspool} improves performance by $140\%$ and achieves the best result in this setting.
These improvements support the hypothesis that aggressive encoder compression is harmful for preserving implicit spatial information in challenging manipulation tasks.
Thus, the limitation comes not from implicit spatial representations themselves, but from applying spatial pooling only after visual features have already been overly compressed.
This motivates {\methodname} to preserve implicit spatial representations across multiple feature scales.

\begin{table}[t]
\centering
\caption{
\textbf{Robomimic PH single-task success rates (\%) over three seeds (150 rollouts).}
{\methodname} is compared with the standard ResNet+{\sspool} baseline.
}
\label{tab:robomimic_main}
\small
\vspace{1mm}
\setlength{\tabcolsep}{5.0pt}
\renewcommand{\arraystretch}{1.12}

\begin{tabular}{llccccc}
\toprule
\textbf{Policy} & \textbf{Encoder}
& \textbf{\textit{Lift}}
& \textbf{\textit{Can}}
& \textbf{\textit{Square}}
& \textbf{\textit{ToolHang}}
& \textbf{Avg.} \\
\midrule

\multirow{3}{*}{\textit{Flow Matching}}
& Base.
& \textbf{99.0}\textpm{}1.0
& 90.0\textpm{}3.0
& 77.0\textpm{}4.0
& 4.7\textpm{}1.2
& 67.7 \\
& {\methodname}
& 96.7\textpm{}4.6
& \textbf{92.7}\textpm{}4.6
& \textbf{84.7}\textpm{}2.3
& \textbf{14.0}\textpm{}4.0
& \textbf{72.0} \\
& \cellcolor{cellgray}Gain
& \cellcolor{cellgray}-2.3
& \cellcolor{cellgray}+2.7
& \cellcolor{cellgray}+7.7
& \cellcolor{cellgray}+9.3
& \cellcolor{cellgray}+4.4\\

\midrule

\multirow{3}{*}{\textit{Diffusion}}
& Base.
& \textbf{97.0}\textpm{}2.0
& 96.0\textpm{}2.0
& 76.0\textpm{}6.0
& 5.3\textpm{}2.3
& 67.1 \\
& {\methodname}
& 96.7\textpm{}3.1
& \textbf{96.7}\textpm{}3.1
& \textbf{77.3}\textpm{}7.0
& \textbf{12.7}\textpm{}8.1
& \textbf{70.9} \\
& \cellcolor{cellgray}Gain
& \cellcolor{cellgray}-0.3
& \cellcolor{cellgray}+0.7
& \cellcolor{cellgray}+1.3
& \cellcolor{cellgray}+7.4
& \cellcolor{cellgray}+3.8 \\

\bottomrule
\end{tabular}
\vspace{-7mm}
\end{table}

\subsection{\RQthree}
We compare {\methodname} with the ResNet+{\sspool} under the same policy architectures and evaluation protocol.
As shown in~\tref{tab:robomimic_main}, {\methodname} improves \textit{ToolHang}, the most geometry-sensitive Robomimic task, from $4.7\%$ to $14.0\%$ under Flow Matching and from $5.3\%$ to $12.7\%$ under Diffusion Policy.
Averaged across both policy backbones, {\methodname} improves \textit{ToolHang} success rate from $5.0\%$ to $13.4\%$, corresponding to a $168.0\%$ relative improvement.
These gains suggest that {\methodname} benefits difficult low-resolution manipulation while maintaining performance on simpler tasks.

These improvements are especially meaningful because \textit{ToolHang} requires precise spatial reasoning over a long manipulation sequence, including grasping, alignment, insertion, and hanging.
By extracting implicit spatial representations from multiple feature scales and aggregating them through top-down cross-attention fusion, {\methodname} provides the policy with richer spatial cues than final-stage {\sspool}.
This addresses the compression bottleneck identified in~\sref{sec:rq2} and supports the gains on geometry-sensitive manipulation.

\begin{wraptable}[8]{r}{0.44\linewidth}
\vspace{-5mm}
\centering
\caption{\textbf{\textit{ToolHang} success rates (\%) across resolution settings (50 rollouts).} }
\label{tab:toolhang_view_resolution}
\vspace{-1mm}
\footnotesize
\setlength{\tabcolsep}{2.6pt}
\renewcommand{\arraystretch}{0.95}
\begin{tabular*}{\linewidth}{@{\extracolsep{\fill}}lcccccc@{}}
\toprule
\multirow{2}{*}{\textbf{Policy}}
& \multicolumn{2}{c}{\textbf{2V-84}}
& \multicolumn{2}{c}{\textbf{1V-240}}
& \multicolumn{2}{c}{\textbf{2V-240}} \\
\cmidrule(lr){2-3}
\cmidrule(lr){4-5}
\cmidrule(l){6-7}
& {Base.} & {\methodname}
& {Base.} & {\methodname}
& {Base.} & {\methodname} \\
\midrule
DP
& 14 & \textbf{26}
& 22 & \textbf{42}
& \textbf{32} & 30 \\
FM
& 16 & \textbf{20}
& 18 & \textbf{30}
& 26 & \textbf{30} \\
MF
& 22 & \textbf{26}
& 16 & \textbf{32}
& 30 & \textbf{44} \\
\bottomrule
\end{tabular*}
\end{wraptable}


We further evaluate whether this improvement extends across different input resolutions and camera views.
As shown in~\tref{tab:toolhang_view_resolution}, {\methodname} improves \textit{ToolHang} performance in most settings across both $84{\times}84$ and $240{\times}240$ observations.
The gains are especially clear for single-view $240{\times}240$ inputs, where {\methodname} improves DP, FM, and MF from $22\%$, $18\%$, and $16\%$ to $42\%$, $30\%$, and $32\%$, respectively.
Although not every setting improves uniformly, these results suggest that preserving multiscale implicit spatial information remains useful across different visual input configurations.

The saliency maps in~\fref{fig:toolhang_saliency} provide qualitative evidence for this improvement.
Compared with the ResNet using other pooling methods, {\methodname} produces cleaner and more localized responses around task-relevant regions across different stages of \textit{ToolHang}.
Across grasping, alignment, insertion, and hanging, {\methodname} consistently attends to the robot arm, the manipulated tool, and their interaction regions.
For example, during the grasp stage, {\methodname} focuses on both the tool and gripper, while other pooling strategies produce more diffuse or less complete responses.
Together with the quantitative results, this saliency analysis suggests that preserving multiscale implicit spatial information helps the visual encoder focus on task-relevant spatial cues for high-precision manipulation.

\subsection{\RQfour}
\begin{figure}[t]
    \centering
    \captionsetup{font=small, labelfont=normalfont}
    \captionsetup[subfigure]{skip=0pt}
    \begin{minipage}[t]{0.75\linewidth}
        \centering
        \vspace{0pt}
        \includegraphics[width=\linewidth]{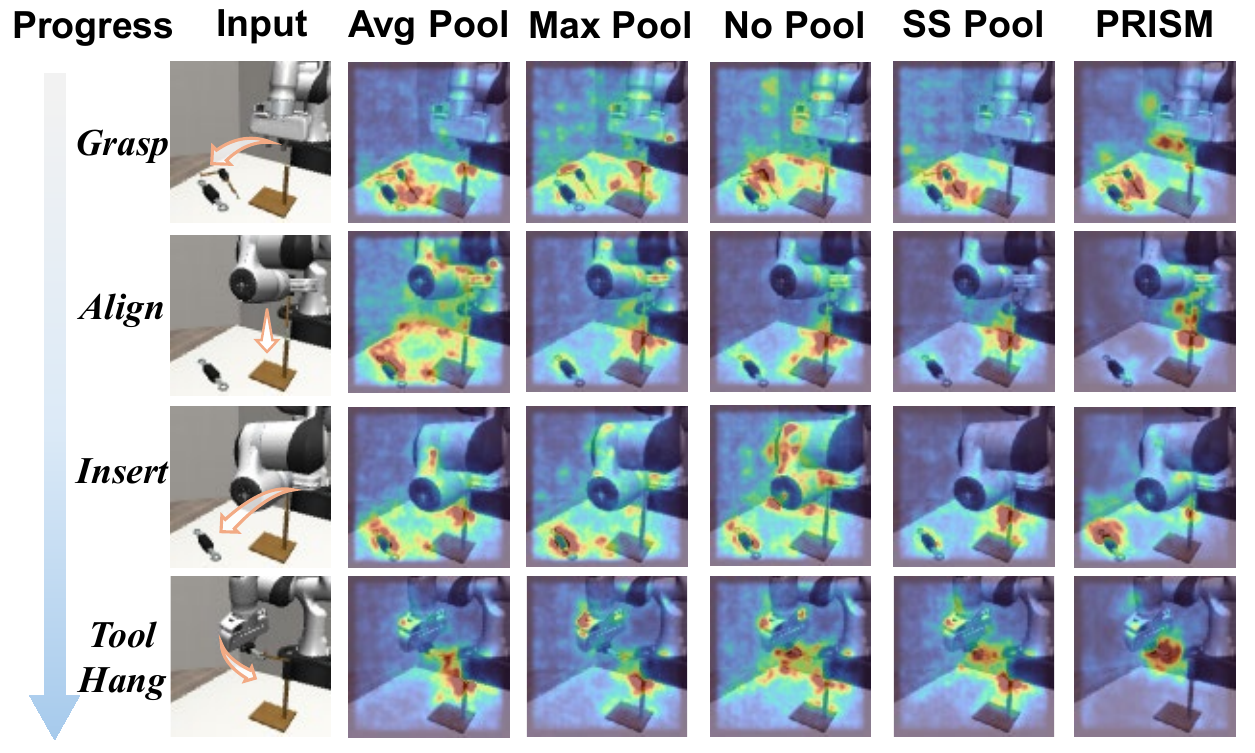}
    \end{minipage}
    \hfill
    \begin{minipage}[t]{0.23\linewidth}
        \centering
        \vspace{0pt}
        \begin{subfigure}[t]{\linewidth}
            \centering
            \includegraphics[width=\linewidth]{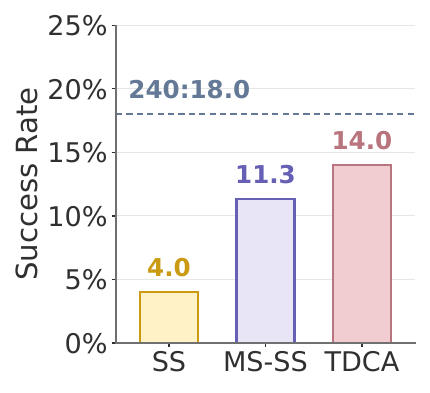}
            \caption{Final SR}
            \label{fig:ablation_raw}
        \end{subfigure}
        \vspace{-1mm}
        \begin{subfigure}[t]{\linewidth}
            \centering
            \includegraphics[width=\linewidth]{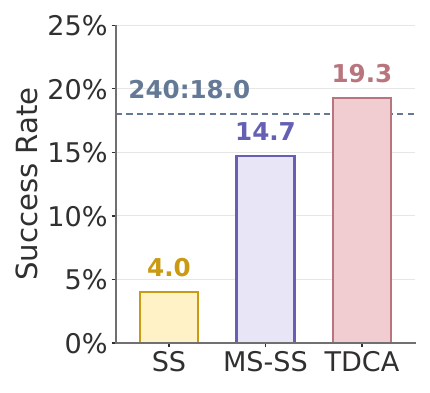}
            \caption{Peak SR}
            \label{fig:ablation_peak}
        \end{subfigure}
    \end{minipage}
    \vspace{1mm}
    \begin{minipage}[t]{0.75\linewidth}
        \captionof{figure}{
        \textbf{Saliency maps on \textit{ToolHang}.} The figures present and compare attended regions across different encoders for the key stages in the \textit{ToolHang} task.
        }
        \label{fig:toolhang_saliency}
    \end{minipage}
    \hfill
    \begin{minipage}[t]{0.23\linewidth}
        \captionof{figure}{
        \textbf{Component ablation on \textit{ToolHang}.}
        }
        \label{fig:ablation_raw_peak}
    \end{minipage}
    \vspace{-8mm}
\end{figure}

To answer this question, we ablate the two main components of {\methodname}: multiscale spatial softmax (MS-SS) and top-down cross-attention fusion (TDCA).
We evaluate them on the single-view $84{\times}84$ \textit{ToolHang} task, using 200 training epochs and 150 rollouts over three seeds.
As shown in~\fref{fig:ablation_raw_peak}, MS-SS substantially improves over final-stage {\sspool}, indicating that preserving implicit spatial information across feature scales is important for high-precision manipulation.
Adding TDCA further improves both the final and peak success rates, showing that top-down fusion helps aggregate multiscale spatial cues more effectively.
Notably, the full model approaches the performance of the $240{\times}240$ reference while using only $84{\times}84$ observations, suggesting that better spatial representation can partially compensate for limited input resolution.

\begin{figure}[t]
\centering

\begin{minipage}[t]{0.68\linewidth}
    \centering
    \vspace{0pt}

    \begin{subfigure}[t]{0.32\linewidth}
        \centering
        \includegraphics[width=\linewidth]{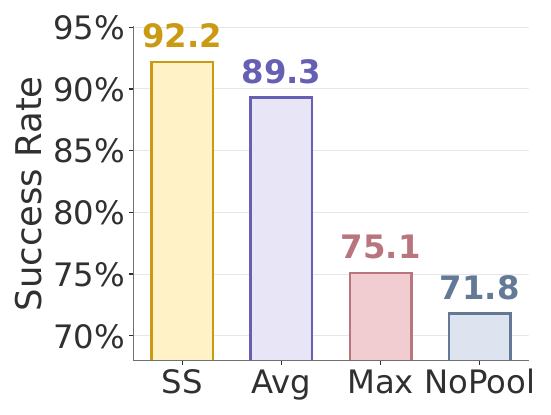}
        \subcaption{Success}
        \label{fig:pool_success}
    \end{subfigure}
    \hfill
    \begin{subfigure}[t]{0.32\linewidth}
        \centering
        \includegraphics[width=\linewidth]{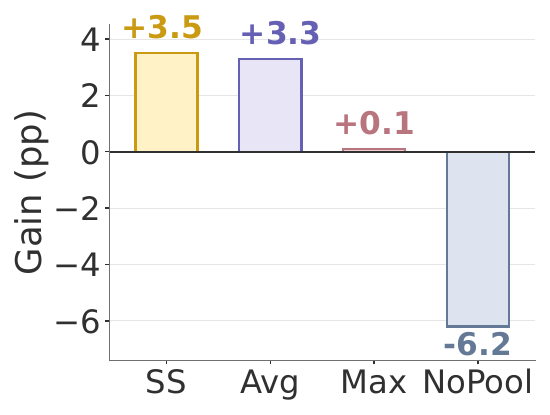}
        \subcaption{Gain}
        \label{fig:pool_gain}
    \end{subfigure}
    \hfill
    \begin{subfigure}[t]{0.32\linewidth}
        \centering
        \includegraphics[width=\linewidth]{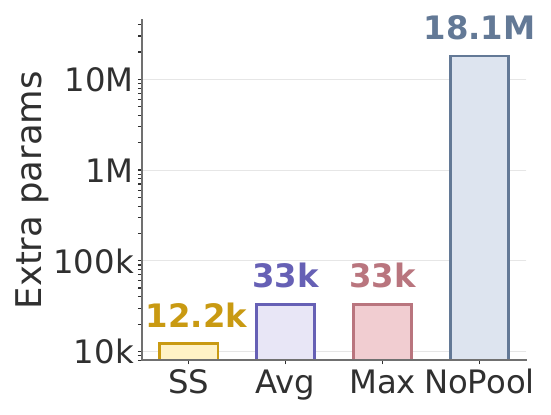}
        \subcaption{Params}
        \label{fig:pool_params}
    \end{subfigure}
    \vspace{-2mm}
    \caption{
    \textbf{Effect of multiscale feature enhancement across pooling operators.}
    The multiscale SSPool mechanism achieves the highest success rate with the lowest parameter cost.
    }
    \label{fig:single_vs_multi_pool}
\end{minipage}
\hfill
\begin{minipage}[t]{0.3\linewidth}
    \centering
    \vspace{-2mm}
    \small
    \setlength{\tabcolsep}{2.5pt}
    \renewcommand{\arraystretch}{1.05}
    \captionsetup{type=table,skip=2pt}

    \captionof{table}{\textbf{LIBERO results.}}
    \label{tab:libero}
    \begin{tabularx}{\linewidth}{@{}lYY@{}}
    \toprule
    \textbf{Encoder} & \textbf{10} & \textbf{Spatial} \\
    \midrule
    Base. & $22.0$ & $37.0$ \\
    {\methodname} & $\mathbf{53.5}$ & $\mathbf{52.5}$ \\
    \bottomrule
    \end{tabularx}

    \vspace{2mm}

    \captionof{table}{\textbf{Real-world results.}}
    \label{tab:realworld_results}
    \begin{tabularx}{\linewidth}{@{}lYY@{}}
    \toprule
    \textbf{Encoder} & \textbf{StackCube} & \textbf{KeyPick} \\
    \midrule
    Base. & $70.0$ & $80.0$ \\
    {\methodname} & $70.0$ & $80.0$ \\
    \bottomrule
    \end{tabularx}
\end{minipage}

\vspace{-6mm}
\end{figure}

We further examine whether the proposed multiscale mechanism benefits other pooling methods.
We evaluate multiscale variants of different pooling operators without the refining blocks on the short-horizon Robomimic tasks and report the results in~\fref{fig:single_vs_multi_pool}.
Multiscale {\sspool} achieves the highest success rate with the smallest parameter overhead, providing the best trade-off between performance and model complexity.
{\avgpool} also benefits from multiscale features, but requires more additional parameters.
In contrast, {\maxpool} yields negligible gain, and {\nopool} degrades performance despite a substantial parameter increase.
These results suggest that multiscale features are most effective when paired with a pooling operator that preserves implicit spatial structure.
We therefore use multiscale {\sspool} as the default spatial extraction module in {\methodname}.

\subsection{\RQfive}
To answer this question, we evaluate {\methodname} beyond single-task Robomimic simulation.
We first conduct multi-task experiments on LIBERO-10~\citep{liu2023libero} and LIBERO-Spatial~\citep{liu2023libero} using the same FM policy as the baseline.
As shown in~\tref{tab:libero}, {\methodname} improves LIBERO-10 from $22.0\%$ to $53.5\%$ and LIBERO-Spatial from $37.0\%$ to $52.5\%$.
These results suggest that preserving multiscale implicit spatial representations also benefits multi-task visuomotor learning.
We further conduct real-world experiments as physical deployment validation.
As shown in~\tref{tab:realworld_results}, {\methodname} successfully completes the StackCube and KeyPick tasks, achieving success rates of $70.0\%$ and $80.0\%$, respectively.
These tasks are relatively short-horizon and less precision-demanding, which may limit the observable benefit of preserving multiscale implicit spatial representations.
Nevertheless, the results show that {\methodname} can be deployed in physical manipulation settings without degrading performance.
Details of the LIBERO dataset and real-world setup are provided in~\aref{app:libero_realworld}.

\section{Limitations}
\label{sec:limitation_discussion}

Although {\methodname} shows consistent relative improvements across policy backbones, tasks, and input resolutions, this work has several limitations.
First, {\methodname} introduces additional computational overhead due to multiscale spatial softmax feature extraction and top-down attention-based fusion.
This overhead is more pronounced during training, since during inference visual features are extracted only once per action chunk.
Second, our real-world experiments are limited in scope.
They mainly serve as deployment validation on relatively simple tasks, where {\methodname} maintains comparable performance to the baseline but does not yet show clear gains.
Future physical experiments should therefore focus on more challenging geometry-sensitive manipulation scenarios.

\section{Discussion}

Despite these limitations, this work provides empirical motivation for choosing implicit spatial representations in robotic manipulation.
Our results suggest that coordinate-aware spatial representations can provide more effective policy inputs than compact feature-value representations.
{\methodname} is one implementation of this principle.
By preserving multiscale implicit spatial information, it addresses the spatial information loss caused by repeated downsampling in standard vision encoders.
Thus, the broader contribution of this work is not only a visual encoder but also evidence that implicit spatial representations are a useful representation principle for visuomotor policy learning.
We do not argue that spatial representations should totally replace semantic representations.
Future work should therefore explore visual encoders that better balance semantic understanding with spatial structure, especially for contact-rich and precision-sensitive manipulation.


\clearpage

\bibliographystyle{assets/plainnat}
\bibliography{example}  

\clearpage
\newpage

\appendix

\section{Appendix}

\subsection{Additional Saliency Map Visualizations}
\label{app:saliency_map}
We provide additional saliency map visualizations to qualitatively analyze how different representation heads affect the visual regions used by the policy.
The saliency maps highlight image regions that have stronger influence on action prediction.

\begin{figure}[h]
    \centering
    \includegraphics[width=\linewidth]{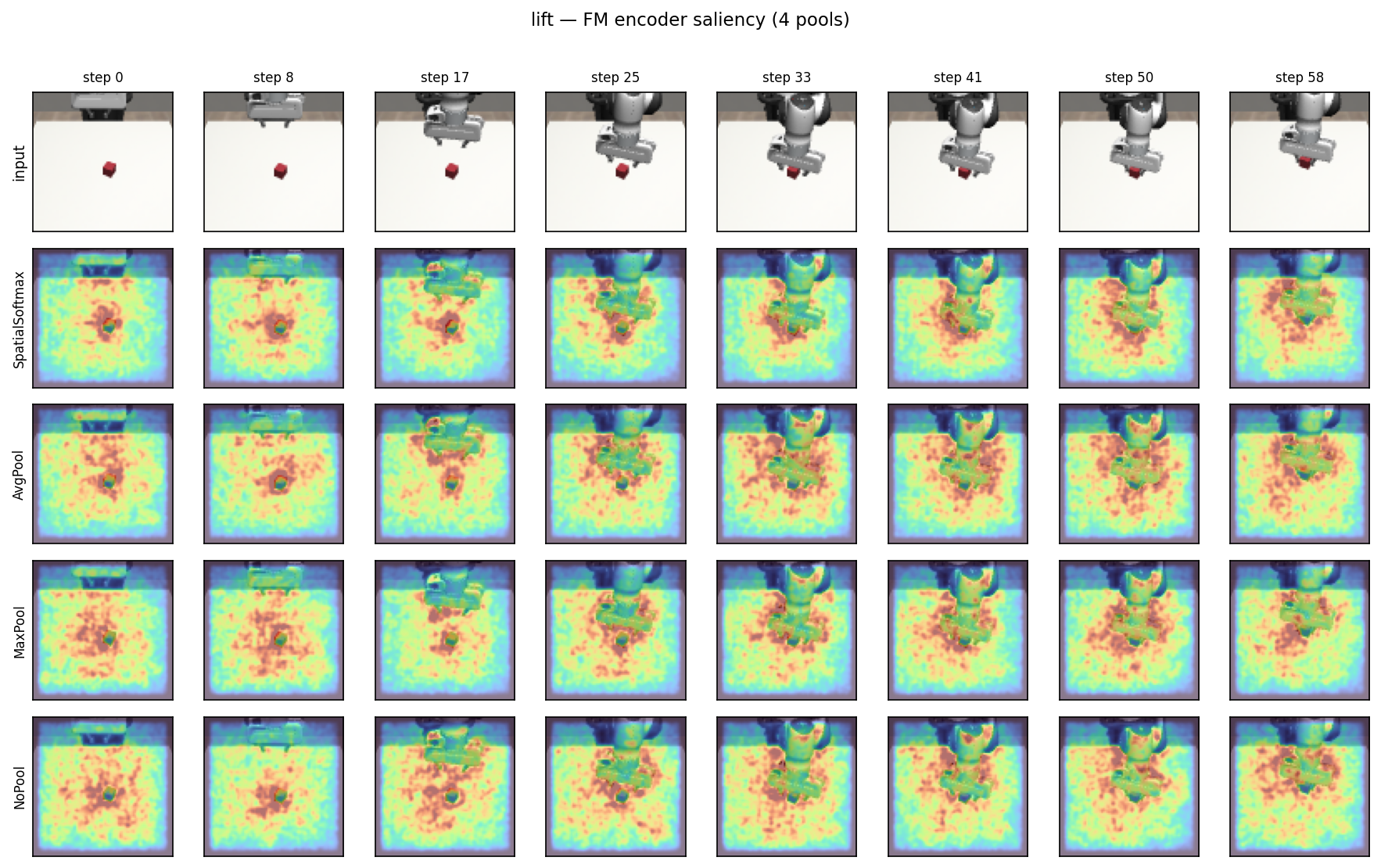}
    \caption{\textbf{Saliency maps with different pooling strategies on the \textit{Lift} task.}
    {\sspool}-based representations tend to produce more localized saliency around the object and gripper than feature-value aggregation.}
    \label{fig:lift_pooling_saliency}
\end{figure}

\begin{figure}
    \centering
    \includegraphics[width=\linewidth]{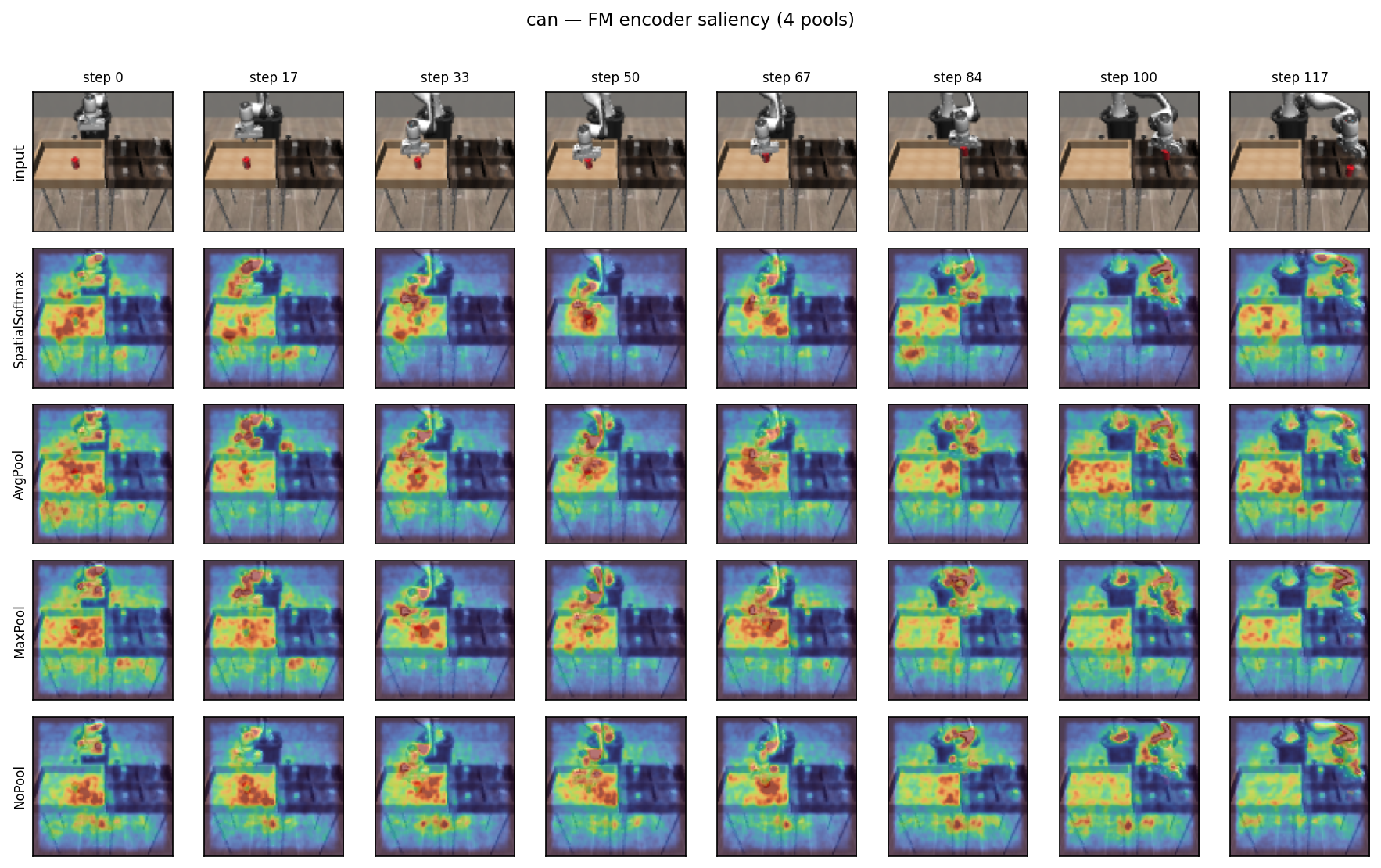}
    \caption{\textbf{Saliency maps with different pooling strategies on the \textit{Can} task.}
    {\sspool}-based representations show more consistent focus on task-relevant object and interaction regions.}
    \label{fig:can_pooling_saliency}
\end{figure}

\begin{figure}
    \centering
    \includegraphics[width=\linewidth]{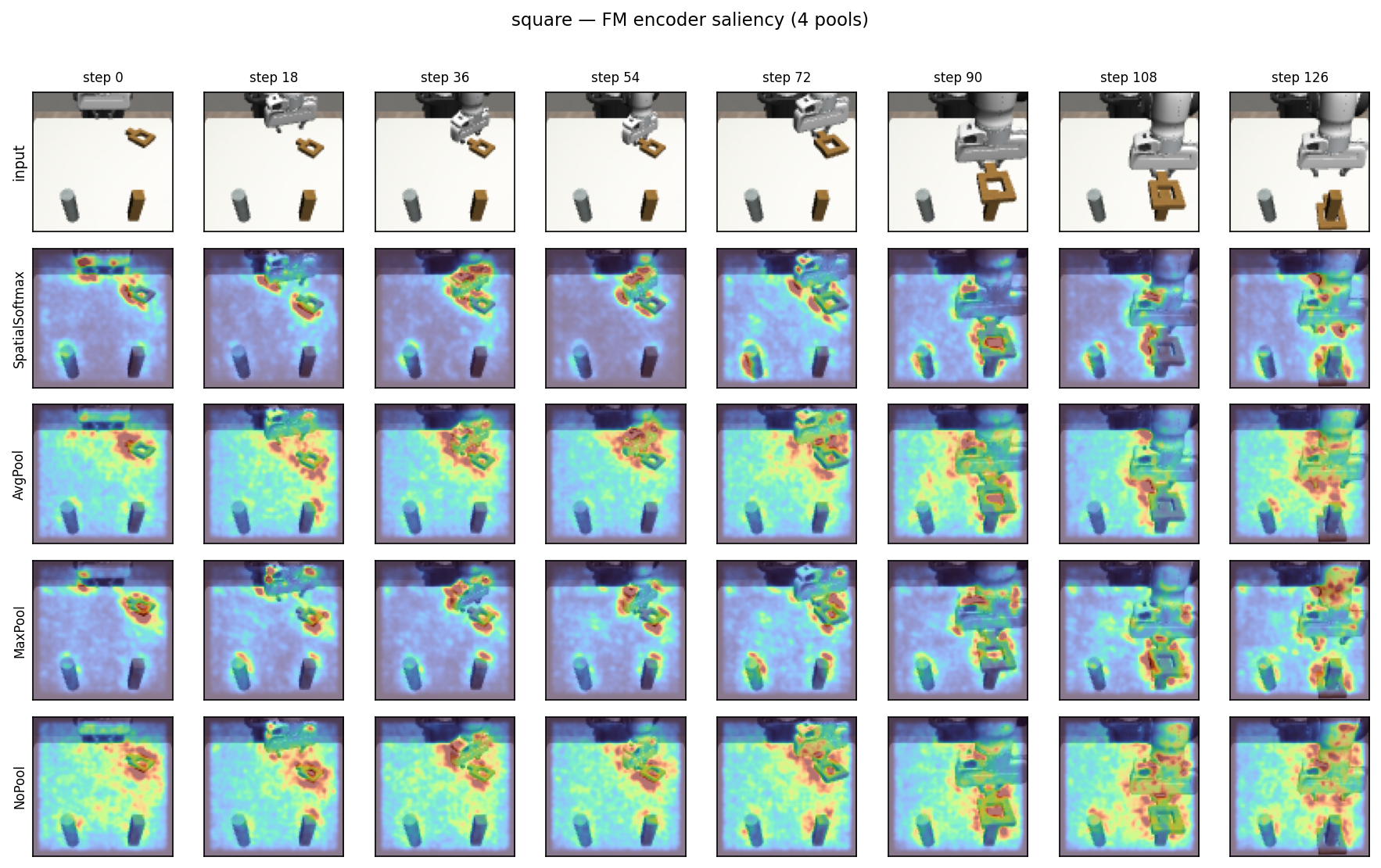}
    \caption{\textbf{Saliency maps with different pooling strategies on the \textit{Square} task.}
    {\sspool}-based representations better preserve spatially localized cues needed for object alignment and placement.}
    \label{fig:square_pooling_saliency}
\end{figure}

We first compare different pooling strategies on the \textit{Lift}, \textit{Can}, and \textit{Square} tasks.
As shown in~\fref{fig:lift_pooling_saliency}, \fref{fig:can_pooling_saliency}, and~\fref{fig:square_pooling_saliency}, {\sspool}-based representations tend to produce more localized and consistent saliency around task-relevant objects, the gripper, and robot-object interaction regions.
In contrast, feature-value aggregation methods, such as average pooling and max pooling, often produce more diffuse or less structured saliency patterns.
These qualitative results are consistent with our quantitative findings, suggesting that implicit spatial representations provide useful policy inputs by exposing where task-relevant visual features occur.

\begin{figure}
    \centering
    \includegraphics[width=\linewidth]{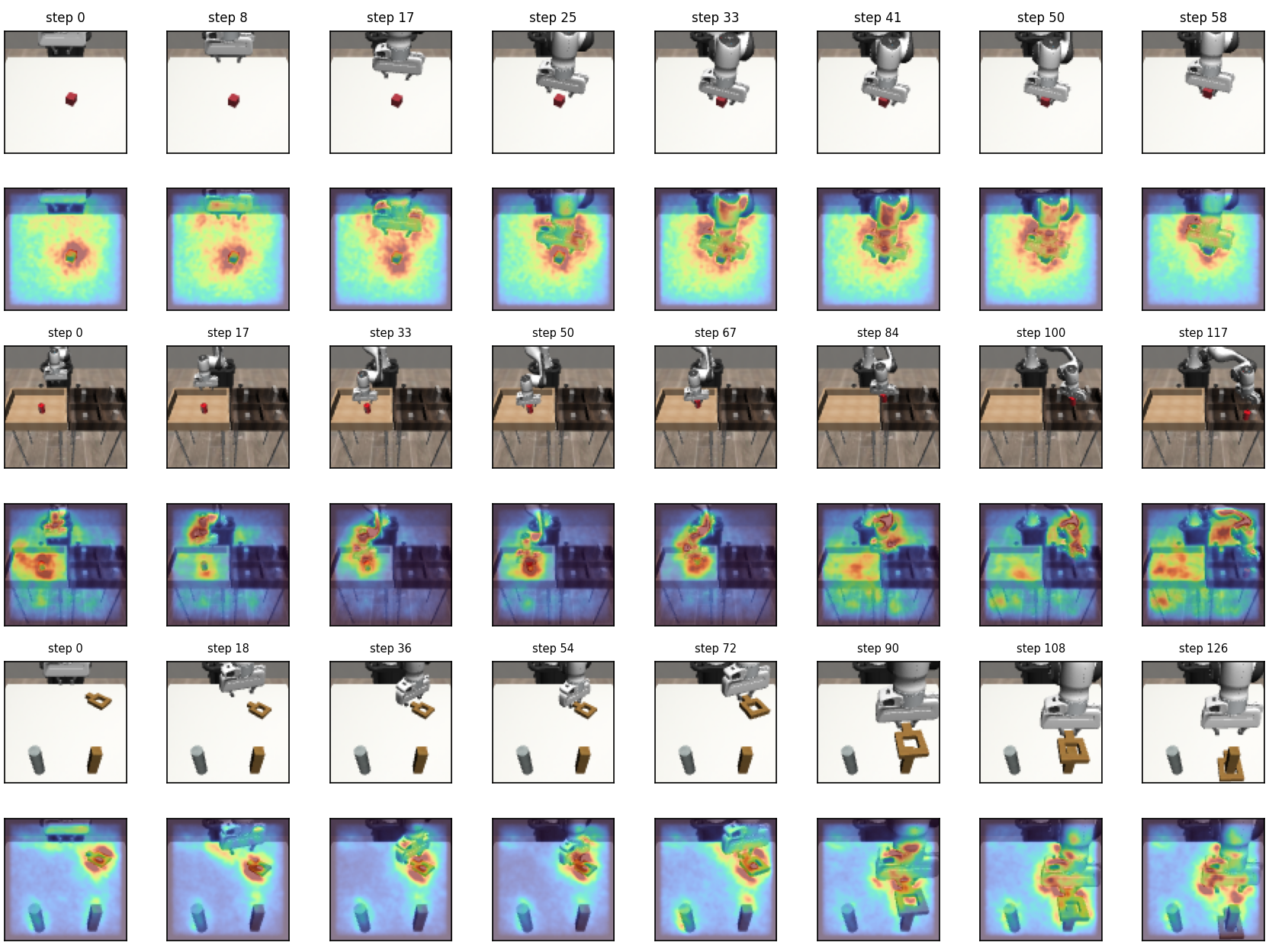}
    \caption{\textbf{Saliency maps of {\methodname} on the \textit{Lift}, \textit{Can}, and \textit{Square} tasks.}
    {\methodname} focuses on task-relevant regions, including the manipulated object, the gripper, and robot-object interaction areas.}
    \label{fig:proposed_saliency_map}
\end{figure}

We further visualize the saliency maps of {\methodname} across the three tasks.
As shown in~\fref{fig:proposed_saliency_map}, {\methodname} focuses on task-relevant regions, including the manipulated object, the gripper, and their interaction areas.
This suggests that preserving multiscale implicit spatial information helps the policy maintain attention on spatially meaningful cues for manipulation.


\newpage
\subsection{Realworld and Multi-Task Evaluation on LIBERO }
\label{app:libero_realworld}

\begin{figure}
    \centering
    \includegraphics[width=\linewidth]{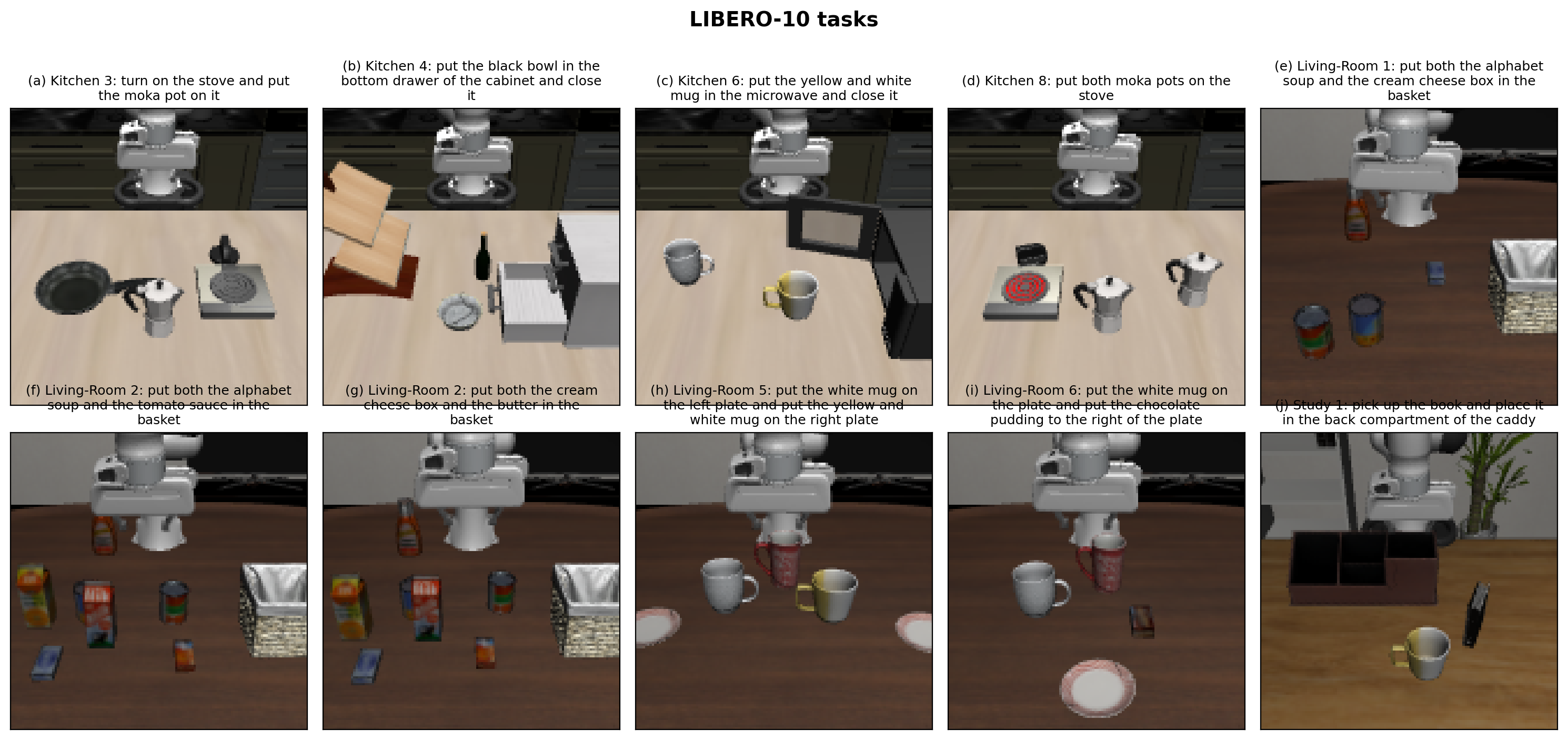}
    \caption{\textbf{Visualization of the LIBERO-10 benchmark tasks.}
    LIBERO-10 contains diverse long-horizon manipulation tasks involving object rearrangement, articulated object interaction, and multi-step language-conditioned control.}
    \label{fig:libero_10_tasks}
\end{figure}

\begin{figure}
    \centering
    \includegraphics[width=\linewidth]{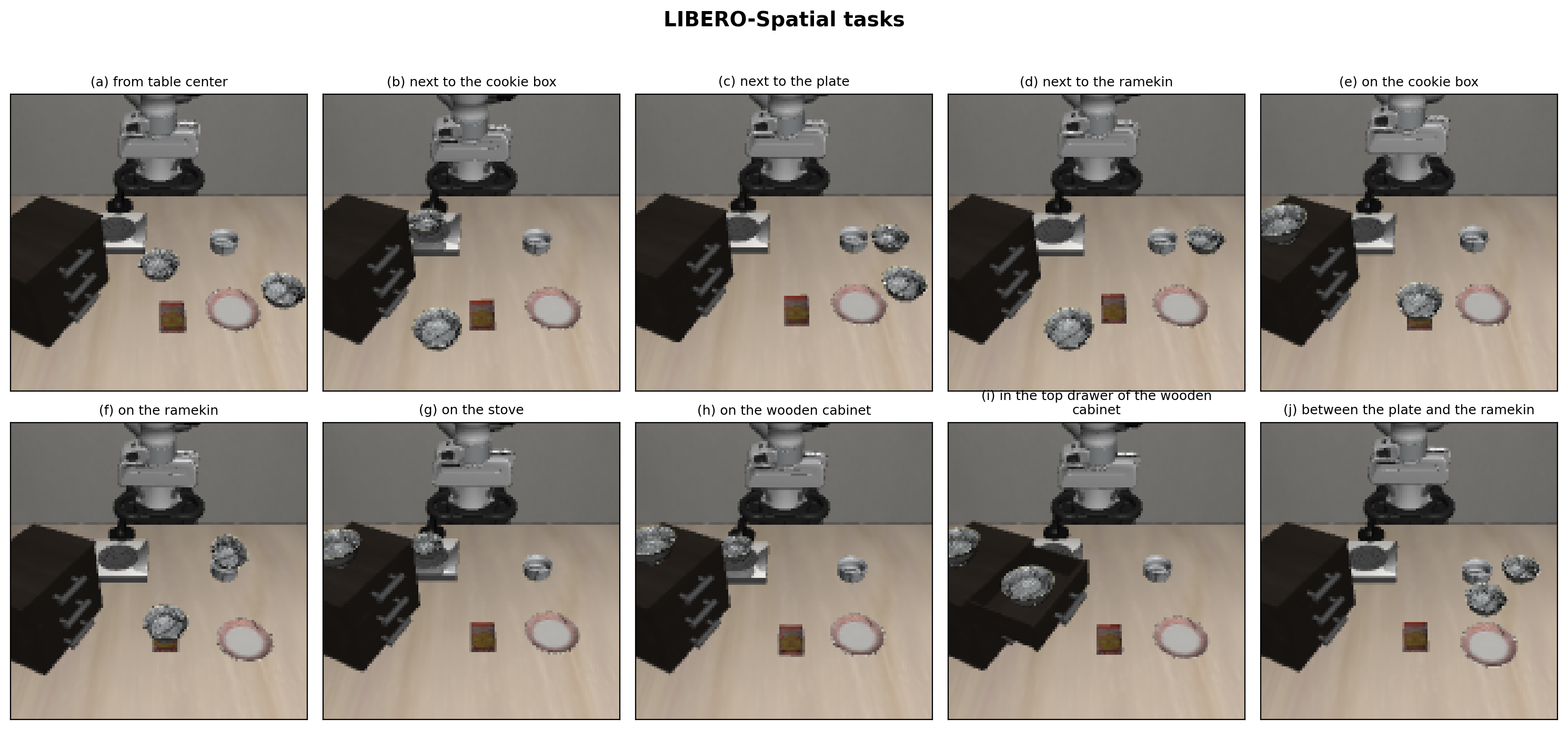}
    \caption{\textbf{Visualization of the LIBERO-Spatial benchmark tasks.}
    LIBERO-Spatial emphasizes spatial reasoning and object arrangement, providing an additional evaluation setting for spatially sensitive manipulation.}
    \label{fig:libero_spatial_tasks}
\end{figure}

We provide additional visualizations of the evaluation settings used in our experiments.
\fref{fig:libero_10_tasks} shows the LIBERO-10 benchmark tasks, which cover diverse long-horizon manipulation scenarios involving object rearrangement, articulated object interaction, and multi-step instruction following.
\fref{fig:libero_spatial_tasks} shows the LIBERO-Spatial tasks, which emphasize spatial reasoning and object arrangement under language-conditioned manipulation.
These tasks provide complementary evaluation settings to Robomimic, allowing us to examine whether preserving implicit spatial representations remains useful beyond the single-task imitation learning setting.

\begin{figure}
    \centering
    \includegraphics[width=\linewidth]{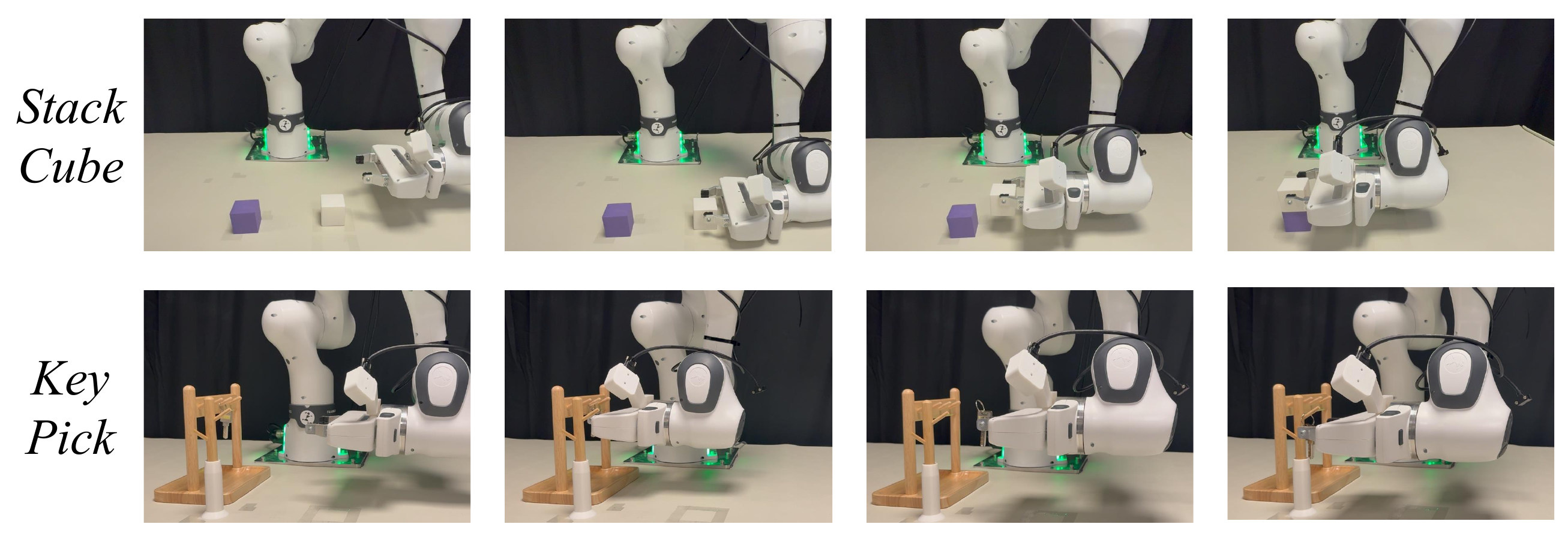}
    \caption{\textbf{Real-world experimental setup.}
    We evaluate the policy on a Franka Panda robot in tabletop manipulation settings that require object localization, robot-object alignment, and contact-aware control.}
    \label{fig:realworld_setup}
\end{figure}

We also visualize the real-world experimental setup in~\fref{fig:realworld_setup}.
The real-world experiments are conducted with a Franka Panda robot under image-based policy control.
The setup includes tabletop manipulation tasks that require accurate object localization, robot-object alignment, and contact-aware motion.
These real-world evaluations provide a practical test of whether the proposed representation design can support spatially precise manipulation beyond simulation.

\newpage
\subsection{Training Details}
\label{app:training_details}

All three baselines (DDPM-UNet, FM-UNet, and MeanFlow-UNet) share the same observation
encoder, UNet backbone, optimizer, and dataloader configuration, differing only in their
noise model and method-specific hyperparameters. All experiments use batch size $512$,
trained for $200$ epochs on Robomimic with image observations of shape $3\times84\times84$.
The following tables document all hyperparameters required to reproduce our results.

\subsubsection{Shared Training Configuration}
\label{app:shared_training}

All methods are trained with AdamW under a cosine learning rate schedule with a linear
warmup period. The encoder is trained end-to-end jointly with the policy network without
any frozen layers. Table~\ref{tab:shared_training} summarises the optimisation
hyperparameters, and Table~\ref{tab:runner} lists the temporal horizon settings shared
across all policies.

\begin{table}[h]
\centering
\small
\begin{tabular}{ll}
\toprule
\textbf{Hyperparameter} & \textbf{Value} \\
\midrule
Batch size (train / val)      & $512$ \\
Number of epochs              & $200$ \\
Steps per epoch               & $250$ \\
Gradient accumulation         & $1$ \\
Optimizer                     & AdamW \\
Learning rate                 & $1\!\times\!10^{-4}$ \\
Betas                         & $(0.9,\,0.999)$ \\
Epsilon                       & $1\!\times\!10^{-8}$ \\
Weight decay                  & $1\!\times\!10^{-6}$ \\
LR schedule                   & Cosine \\
LR warmup steps               & $500$ \\
Encoder                       & Trained end-to-end (not frozen) \\
Rollout / checkpoint interval & $50$ epochs \\
\bottomrule
\end{tabular}
\caption{Shared optimisation hyperparameters used across all methods.}
\label{tab:shared_training}
\end{table}

\begin{table}[h]
\centering
\small
\begin{tabular}{ll}
\toprule
\textbf{Hyperparameter} & \textbf{Value} \\
\midrule
Prediction horizon $H$ & $16$ \\
Observation steps $T_o$ & $2$ \\
Action steps $T_a$      & $8$ \\
Image shape             & $3\times84\times84$ \\
\bottomrule
\end{tabular}
\caption{Shared runner configuration. $H$ denotes the total prediction horizon; $T_o$ and
$T_a$ are the number of observation context frames and executed action steps per rollout
cycle, respectively.}
\label{tab:runner}
\end{table}

\subsubsection{Shared Observation Encoder}
\label{app:encoder}

All methods use a \texttt{MultiImageObsEncoder} that processes each camera view
independently through a ResNet-18 backbone initialised from scratch (no ImageNet
pretraining). BatchNorm layers are replaced by GroupNorm to improve stability under
the small effective per-device batch sizes that arise with image observations. Random
cropping and ImageNet-style mean/std normalisation are applied as data augmentation
during training. Camera-specific weights are \emph{not} shared, allowing the encoder
to specialise to each viewpoint. The full configuration is given in
Table~\ref{tab:encoder}.

\begin{table}[h]
\centering
\small
\begin{tabular}{ll}
\toprule
\textbf{Hyperparameter} & \textbf{Value} \\
\midrule
Encoder                        & \texttt{MultiImageObsEncoder} \\
RGB backbone                   & ResNet-18 (random init, no pretraining) \\
Normalization                  & GroupNorm (BatchNorm replaced) \\
Random crop                    & Enabled \\
ImageNet normalization         & Enabled \\
Share RGB model across cameras & False \\
\bottomrule
\end{tabular}
\caption{Shared observation encoder configuration. GroupNorm is used in place of
BatchNorm throughout the ResNet-18 backbone.}
\label{tab:encoder}
\end{table}

\subsubsection{Shared 1D Conditional UNet}
\label{app:unet}

The denoising backbone is a 1D temporal UNet that conditions on the encoded observation
via FiLM-style modulation (\texttt{cond\_predict\_scale=True}). The network operates on
action sequences of length $H=16$ and uses three resolution levels with channel widths
$[256, 512, 1024]$. Diffusion timestep embeddings of dimension $128$ are injected at
each resolution level. This backbone is shared identically across DDPM-UNet, FM-UNet,
and MeanFlow-UNet; all three methods therefore have the same parameter count and
receptive field. The configuration is listed in Table~\ref{tab:unet}.

\begin{table}[h]
\centering
\small
\begin{tabular}{ll}
\toprule
\textbf{Hyperparameter} & \textbf{Value} \\
\midrule
Diffusion-step embedding dim  & $128$ \\
Down channels                 & $[256,\,512,\,1024]$ \\
Kernel size                   & $5$ \\
Group-norm groups             & $8$ \\
\texttt{cond\_predict\_scale} & True \\
\bottomrule
\end{tabular}
\caption{Shared 1D conditional UNet backbone used by all three methods. The same
architecture serves as the denoising network (DDPM-UNet), velocity network (FM-UNet),
and mean-flow network (MeanFlow-UNet).}
\label{tab:unet}
\end{table}

\subsubsection{Method-Specific Hyperparameters}
\label{app:method_specific}

While the backbone and training recipe are shared, each method introduces its own
noise process and inference procedure. Table~\ref{tab:method_configs} summarises
these differences. DDPM-UNet follows the standard DDPM formulation with a squared
cosine schedule and $\epsilon$-prediction. FM-UNet uses a flow matching objective
with a straight-path probability flow and requires $10$ Euler steps at inference.
MeanFlow-UNet replaces the multi-step ODE solver with a single-step mean-flow
estimator, trading a richer training recipe (logit-normal time sampling, adaptive
loss weighting, EMA) for zero-shot one-step inference.

\begin{table}[h]
\centering
\small

\begin{subtable}[t]{0.48\linewidth}
\centering
\begin{tabular}{ll}
\toprule
\textbf{Hyperparameter} & \textbf{Value} \\
\midrule
Noise scheduler       & \texttt{DDPMScheduler} \\
$\beta$ start         & $1\!\times\!10^{-4}$ \\
$\beta$ end           & $2\!\times\!10^{-2}$ \\
$\beta$ schedule      & \texttt{squaredcos\_cap\_v2} \\
Variance type         & \texttt{fixed\_small} \\
Train timesteps       & $100$ \\
Inference steps       & $100$ \\
Prediction type       & $\epsilon$ \\
\texttt{clip\_sample} & True \\
\bottomrule
\end{tabular}
\caption{DDPM-UNet}
\label{tab:ddpm}
\end{subtable}
\hfill
\begin{subtable}[t]{0.48\linewidth}
\centering
\begin{tabular}{ll}
\toprule
\textbf{Hyperparameter} & \textbf{Value} \\
\midrule
Inference steps               & $10$ \\
\bottomrule
\end{tabular}
\caption{FM-UNet}
\label{tab:fm}
\end{subtable}

\vspace{1em}

\begin{subtable}[t]{\linewidth}
\centering
\begin{tabular}{ll}
\toprule
\textbf{Hyperparameter} & \textbf{Value} \\
\midrule
Noise scheduler                                  & \texttt{DDPMScheduler} (train timesteps $=100$) \\
$\beta$ start / end                              & $1\!\times\!10^{-4}$ / $2\!\times\!10^{-2}$ \\
$\beta$ schedule                                 & \texttt{squaredcos\_cap\_v2} \\
Adaptive-weight norm power $p$                   & $0.75$ \\
Adaptive-weight $\epsilon$                       & $1\!\times\!10^{-3}$ \\
EDM augmentation                                 & Disabled \\
EMA decay                                        & $0.9999$ \\
Time sampler                                     & \texttt{v1} (logit-normal) \\
$P_{\mathrm{mean}}^{t}$, $P_{\mathrm{std}}^{t}$ & $-0.6$, $1.6$ \\
$P_{\mathrm{mean}}^{r}$, $P_{\mathrm{std}}^{r}$ & $-4.0$, $1.6$ \\
$t{=}r$ sampling ratio                           & $0.75$ \\
Inference                                        & One-step \\
\bottomrule
\end{tabular}
\caption{MeanFlow-UNet}
\label{tab:meanflow}
\end{subtable}

\caption{Method-specific hyperparameter configurations. \textbf{(a)}~DDPM-UNet uses
a squared-cosine noise schedule with $\epsilon$-prediction and $100$ DDPM sampling
steps. \textbf{(b)}~FM-UNet uses a flow matching loss with straight probability paths
and $10$ Euler integration steps at inference. \textbf{(c)}~MeanFlow-UNet trains a
mean-flow estimator with logit-normal time sampling and adaptive loss weighting,
enabling single-step inference at test time.}
\label{tab:method_configs}
\end{table}

\newpage

\subsection{Training Objective}
\label{app:objectives}

This appendix details the per-sample training objectives of the three baselines
used in our experiments. All three policies share the same observation encoder
and 1D conditional UNet backbone, so the descriptions below isolate the
generative objective itself.

\paragraph{Notation.}
Let $\tau \in \mathbb{R}^{H \times d_a}$ denote a normalized action chunk of
horizon $H$, and let $c = \mathrm{Enc}_{\psi}(\mathbf{o}_{1:T_o})$ denote the
global condition produced by the observation encoder from the past $T_o$
observation frames. Each policy parameterizes a UNet $f_\theta(\cdot, \cdot, c)$
that operates on action chunks; we use $M$ to denote the action loss-mask
(complement of the inpainting / conditioning mask). Expectations
$\mathbb{E}[\cdot]$ are taken over the data distribution of $(\tau, c)$ and over
all stochastic variables introduced within each objective.

\subsection{Denoising Diffusion (DDPM) Policy}

We adopt the standard $\epsilon$-prediction DDPM formulation~\citep{ho2020denoising}
with $K=100$ training timesteps and a squared-cosine $\bar\alpha$ schedule. For
each sample, a diffusion step $k \sim \mathcal{U}\{0,\ldots,K-1\}$ and a Gaussian
perturbation $\epsilon \sim \mathcal{N}(\mathbf{0}, \mathbf{I})$ are drawn, and
the noisy trajectory is constructed as
\begin{equation}
\tau_k \;=\; \sqrt{\bar\alpha_k}\,\tau \;+\; \sqrt{1-\bar\alpha_k}\,\epsilon.
\end{equation}
The network is trained to predict the injected noise by minimizing the masked
mean-squared error
\begin{equation}
\mathcal{L}_{\mathrm{DDPM}}(\theta,\psi)
\;=\; \mathbb{E}\!\left[
\bigl\lVert\, M \odot \bigl(\epsilon_\theta(\tau_k, k, c) - \epsilon \bigr) \bigr\rVert_2^2
\right].
\end{equation}
At inference, action chunks are generated by running the DDPM ancestral sampler
for $K=100$ denoising steps using exponential-moving-average (EMA) weights of
$\theta$ and $\psi$.

\subsection{Flow Matching (FM) Policy}

We adopt conditional flow matching~\citep{lipman2023flow,liu2023flow} along
a straight (rectified) probability path between a Gaussian prior and the data
distribution. For each sample, a prior point
$x_0 \sim \mathcal{N}(\mathbf{0}, \mathbf{I})$ and a time
$t \sim \mathcal{U}(0, 1)$ are drawn, yielding the interpolant
\begin{equation}
x_t \;=\; (1-t)\,x_0 \;+\; t\,\tau,
\qquad
u^{\star}(x_t \mid \tau, x_0) \;=\; \tau - x_0,
\end{equation}
where $u^{\star}$ is the constant ground-truth velocity of the straight path.
The network $v_\theta(x_t, t, c)$ regresses this velocity under the masked
squared loss
\begin{equation}
\mathcal{L}_{\mathrm{FM}}(\theta,\psi)
\;=\; \mathbb{E}\!\left[
\bigl\lVert\, M \odot \bigl(v_\theta(x_t, t, c) - (\tau - x_0) \bigr) \bigr\rVert_2^2
\right].
\end{equation}
Action chunks are obtained at inference by Euler-integrating
$\dot{x}_t = v_\theta(x_t, t, c)$ from
$x_0 \sim \mathcal{N}(\mathbf{0}, \mathbf{I})$ at $t=0$ to $x_1 = \tau$ at $t=1$
in $10$ uniform steps with EMA weights. Throughout this paper the auxiliary
state-consistency terms exposed by the implementation are disabled
($w_{\mathrm{FM}}=1$ and all auxiliary weights are zero).

\subsection{MeanFlow Policy}

MeanFlow~\citep{geng2026mean} learns the \emph{average} velocity field of the
same straight probability path between two times $r \le t$, so that an arbitrary
flow segment can be traversed in a single network evaluation. Defining
\begin{equation}
\bar{u}^{\star}(z, r, t)
\;=\; \frac{1}{t-r}\int_{r}^{t} u^{\star}(z_s, s)\,\mathrm{d}s,
\end{equation}
the average velocity satisfies the MeanFlow identity
\begin{equation}
\bar{u}^{\star}(z, r, t)
\;=\; u^{\star}(z, t) \;-\; (t-r)\,\tfrac{\mathrm{d}}{\mathrm{d}t}\,\bar{u}^{\star}(z, r, t),
\label{eq:mf_identity}
\end{equation}
which provides a self-consistent regression target without requiring numerical
integration. We parameterize the average velocity as $u_\theta(z, t, h, c)$ with
$h = t - r$ and train it using a single network call together with one
Jacobian-vector product (JVP), as follows.

For each sample, an independent Gaussian perturbation
$e \sim \mathcal{N}(\mathbf{0}, \mathbf{I})$ and a pair of times $(t, r)$ are
drawn. The two times are sampled from independent logit-normal distributions
(parameters
$P^{t}_{\mathrm{mean}}=-0.6,\;P^{t}_{\mathrm{std}}=1.6$,
$P^{r}_{\mathrm{mean}}=-4.0,\;P^{r}_{\mathrm{std}}=1.6$), and with probability
$0.75$ we set $r = t$ to bias training toward the instantaneous-velocity
boundary case. The interpolant and instantaneous velocity along the straight
path are
\begin{equation}
z \;=\; (1-t)\,\tau \;+\; t\,e,
\qquad
v \;=\; e - \tau.
\end{equation}
The total derivative of $u_\theta$ along the path is then computed by a JVP
along the tangent $(\dot{z}, \dot{t}, \dot{r}) = (v, 1, 0)$:
\begin{equation}
\bigl(u_{\mathrm{pred}},\; \dot{u}_\theta\bigr)
\;=\; \mathrm{JVP}\bigl(u_\theta;\; (z, t, r),\; (v, 1, 0)\bigr).
\end{equation}
Substituting into~\eqref{eq:mf_identity} yields the stop-gradient regression
target
\begin{equation}
u_{\mathrm{tgt}}
\;=\; \mathrm{sg}\!\left[\, v \;-\; (t - r)\,\dot{u}_\theta \,\right].
\end{equation}
We train with the adaptive $L_2$ loss proposed
by~\citet{geng2026mean}: with
$\Delta = \lVert u_{\mathrm{pred}} - u_{\mathrm{tgt}} \rVert_2^2$,
\begin{equation}
\mathcal{L}_{\mathrm{MF}}(\theta,\psi)
\;=\; \mathbb{E}\!\left[
\frac{\Delta}{\bigl(\mathrm{sg}[\Delta] + \varepsilon\bigr)^{p}}
\right],
\qquad p = 0.75,\quad \varepsilon = 10^{-3},
\end{equation}
which down-weights large residuals and stabilizes the JVP-based target. At
inference, a single forward pass with $t = 1$ and $r = 0$ produces the action
chunk $\tau = z - u_\theta(z, 1, 1, c)$ from $z \sim \mathcal{N}(\mathbf{0},\mathbf{I})$,
i.e.\ one-step generation. EMA weights of decay $0.9999$ are used during
evaluation.

\label{app:libero_real}
\end{document}